\documentclass[conference]{IEEEtran}
\renewcommand{\footnoterule}{%
    \kern -3pt 
    \hrule width 0.4\linewidth height 0.4pt 
    \kern 2pt 
}
\IEEEoverridecommandlockouts
\usepackage{cite}
\usepackage{amsmath,amssymb,amsfonts}
\usepackage{algorithmic}
\usepackage{graphicx}
\usepackage{textcomp}
\usepackage{xcolor}
\usepackage{booktabs}
\usepackage{multicol}
\usepackage{multirow}
\usepackage{pifont}
\def\BibTeX{{\rm B\kern-.05em{\sc i\kern-.025em b}\kern-.08em
    T\kern-.1667em\lower.7ex\hbox{E}\kern-.125emX}}
\begin{document}

\title{Target-oriented Multimodal Sentiment Classification with Counterfactual-enhanced Debiasing}

\author{\IEEEauthorblockN{Zhiyue Liu$^{1,2}$, Fanrong Ma$^{1}$, Xin Ling$^{3*}$}
\IEEEauthorblockA{$^1$School of Computer, Electronics and Information, Guangxi University, Nanning, China}
\IEEEauthorblockA{$^2$Guangxi Key Laboratory of Multimedia Communications and Network Technology}
\IEEEauthorblockA{$^3$School of Sociology and Anthropology, Sun Yat-sen University, Guangzhou, China}
\IEEEauthorblockA{liuzhy@gxu.edu.cn, 2213301037@st.gxu.edu.cn, lingx29@mail2.sysu.edu.cn}}

\maketitle

\renewcommand{\thefootnote}{}
\footnotetext{$^*$Corresponding author.}

\begin{abstract}
Target-oriented multimodal sentiment classification seeks to predict sentiment polarity for specific targets from image-text pairs. While existing works achieve competitive performance, they often over-rely on textual content and fail to consider dataset biases, in particular word-level contextual biases. This leads to spurious correlations between text features and output labels, impairing classification accuracy. In this paper, we introduce a novel counterfactual-enhanced debiasing framework to reduce such spurious correlations. Our framework incorporates a counterfactual data augmentation strategy that minimally alters sentiment-related causal features, generating detail-matched image-text samples to guide the model's attention toward content tied to sentiment. Furthermore, for learning robust features from counterfactual data and prompting model decisions, we introduce an adaptive debiasing contrastive learning mechanism, which effectively mitigates the influence of biased words. Experimental results on several benchmark datasets show that our proposed method outperforms state-of-the-art baselines.

\end{abstract}


\section{Introduction}
\label{sec:intro}

With the rise of multimedia social platforms like Twitter, users have increasingly preferred expressing their feelings and emotions about various events or topics through multimodal posts. As a result, there has been growing interest in multimodal sentiment analysis. Among these, Target-oriented multimodal sentiment classification (TMSC) is a fine-grained task in sentiment analysis, which aims to predict the sentiment polarity of a target entity mentioned in the textual modality of a given image-text pair.

Most TMSC methods address the task as an input-output mapping problem based on deep neural networks. Some early works~\cite{yu2019adapting,yang2024prompt} aim to design attention mechanisms or follow the pre-training and fine-tuning paradigm to capture the interactions among the text, image, and target. To achieve alignment in different modal feature spaces,~\cite{khan2021exploiting,wang2024image}
propose a transformer-based image captioning model that translates images into textual modality. Another line of studies~\cite{xiao2022adaptive,wang2024dual} utilizes syntactic knowledge and graph neural networks to fuse information from the text and image.

\begin{figure}[t]
\centering 
\includegraphics[scale=0.42]{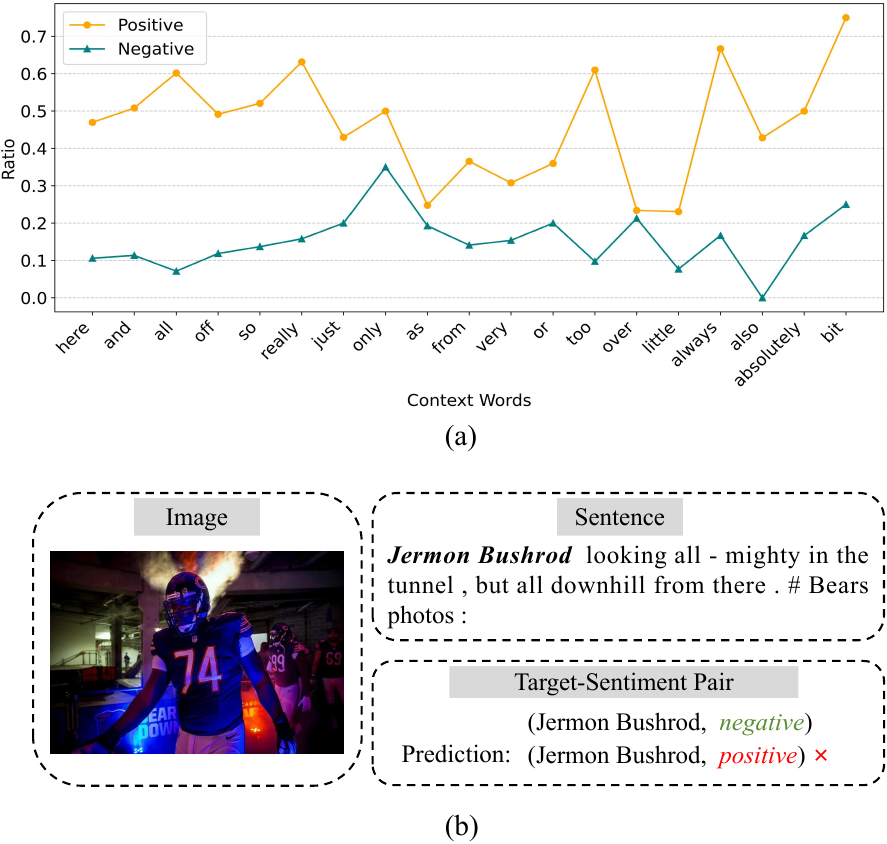} 
\caption{Several context words from the training set of benchmark datasets and an example of TMSC.}
\label{fig1}
\end{figure}

Although prior work has made progress, few of them focus on addressing the issue of spurious correlations. That is, neural networks trained with the stochastic gradient descent algorithm are prone to biases introduced by the dataset~\cite{wu2024diner}. As a result, the model associates certain irrelevant input features with output labels, learning shortcuts instead of the underlying task. The TMSC models also suffer from biases in the dataset, particularly word-level contextual biases within the text. Figure~\ref{fig1} (a) exhibits that certain words unrelated to sentiment expression exhibit imbalanced frequencies between negative and positive samples in benchmark datasets. Since textual information dominates the fusion decision process and plays a crucial role in addressing TMSC~\cite{ye2023rethinking}, biases from texts impact sentiment classification. This causes the model to rely on biased statistical patterns rather than the inherent semantics of the text to predict samples containing biased words. As shown in Figure~\ref{fig1} (b), contextual words like ``all" and ``from" in the sentence have a high co-occurrence rate with the positive label, which may cause existing methods~\cite{yu2019adapting,wang2024image,wang2024dual} to make unreliable predictions as ``positive".

The influence of biased statistical information constrains the accuracy of sentiment classification. Thus, we propose a novel counterfactual-enhanced debiasing framework (CED) to mitigate such spurious correlations. Specifically, we design a counterfactual data augmentation strategy tailored for the TMSC task to make the model focus more on sentiment-related content. This strategy generates counterfactual samples from the original training data by minimally modifying the sentiment-related causal features. Existing text-only counterfactual augmentation methods are unsuitable for our scenario, as sentiment expressions in images and text are often closely interconnected. For example, as shown in Figure~\ref{fig1}~(b), the dim background in the image conveys distinct negative emotional cues. Relying solely on text may lead to an erroneous prediction for the ``Jermon Bushrod'' target words. Therefore, while editing the textual content, we also edit the details of the images to obtain sentiment-consistent image-text pairs.
Furthermore, we design an adaptive debiasing contrastive learning mechanism to distinguish negative samples containing similar context-biased words while reducing the gap between positive samples. A re-weighted contrastive loss function is employed to push apart the feature representations of negative samples with similar biased words but opposite labels while pulling closer the feature representations of samples with the same label. This allows the model to focus on meaningful multimodal semantic sentiment cues specific to the target words, thereby improving classification accuracy. Our main contributions are summarized as follows:

\begin{itemize}
\item  A novel CED framework for TMSC is proposed to reduce spurious correlations through counterfactual data augmentation and adaptive debiasing contrastive learning, which enables the learning of a more robust model.
\item A counterfactual data augmentation strategy for image-text pairs is designed by altering sentiment-related causal features subtly, which guides the model to focus on sentiment content and improves the classification accuracy.
\item Experimental results on the Twitter benchmark dataset~\cite{yu2019adapting} demonstrate that our proposed method outperforms state-of-the-art baselines.
\end{itemize}

\section{Related Work}
\subsection{Target-oriented Multimodal Sentiment Classification}
TMSC is a crucial fine-grained task in multimodal sentiment analysis. Deep learning methods have dominated this field. Early approaches rely primarily on attention mechanisms or adapt existing pre-trained models to facilitate cross-modal interactions. \cite{yu2019adapting} first exploit BERT~\cite{devlin2018bert} and ResNet~\cite{he2016deep} as the backbone to extract and fuse features of textual and visual modalities. A multimodal prompt fusion interaction Transformer is proposed by~\cite{yang2024prompt}, which enhances sentiment analysis by refining texture features and focusing on sentiment-relevant information. Considering the problem of multimodal feature fusion across diverse feature spaces, some studies have transformed visual information into a linguistic format. \cite{khan2021exploiting} combines the original and translated image captions to predict sentiment polarity.~\cite{xiao2022adaptive} proposes a two-stream adaptive multi-feature extraction graph convolutional network to fuse semantic and syntactic features from sentences and image captions. Recently, instruction learning~\cite{yang2023visual,feng2024a2II} has been used to address the effective fusion of text and image information. However, these studies mainly rely on the text modality~\cite{ye2023rethinking}, overlooking the impact of word-level contextual biases.

\subsection{Counterfactual Data Augmentation in Sentiment Analysis}
Augmenting training datasets with counterfactual examples could boost the robustness of sentiment analysis and natural language inference classifiers~\cite{kaushik2020learning}. Therefore, in the domain of text-only sentiment analysis, different counterfactual augmentation strategies are designed to mitigate issues of spurious correlations~\cite{chang2024counterfactual}. For instance,~\cite{wang2021robustness} proposes an approach to automatically construct counterfactual samples by replacing or removing potential causal words in the original training data. Diffusion-based counterfactual data augmentation could generate high-quality and diverse samples~\cite{xin2024diffusion}.
Despite significant progress in text-only tasks, the application of counterfactual data in the multimodal domain emerges slowly.~\cite{jia2024debiasing} initially employs ChatGPT to rewrite only textual information in the sarcastic image-text sample. However, for the TMSC task, data augmentation involves not only modifying the text but also processing the images to ensure consistency between the sentiment details conveyed by the image-text pair.

\section{Methodology}
Our CED framework includes two components: a counterfactual data augmentation strategy and an adaptive contrastive learning mechanism. To enhance the model's sensitivity to sentiment-related features in multimodal data, we make minimal modifications to the causal features related to sentiment in image-text pairs. For learning robust features, we assign different weights to sample pairs based on feature distances, pulling apart samples with similar biased words but different labels while bringing closer those with the same labels.

\subsection{Task Definition}
Given a dataset $\mathcal{D}$ including image-text pairs, each pair consists of a sentence $S = (w_1, ..., w_n)$ with $n$ words, an image $V$, and a target $T = (t_1, ..., t_m)$ with $m$ words, where the target $T$ is a subsequence of $S$ and labeled with a sentiment polarity $y \in  \{\rm{negative, neutral, positive}\}$. The goal of TMSC is to learn a target-oriented classifier and predict the sentiment label of targets by using the sentence and associated image.
\subsection{Counterfactual Multimodal Data Augmentation}
\begin{figure}[t]
\centering 
\includegraphics[scale=0.34]{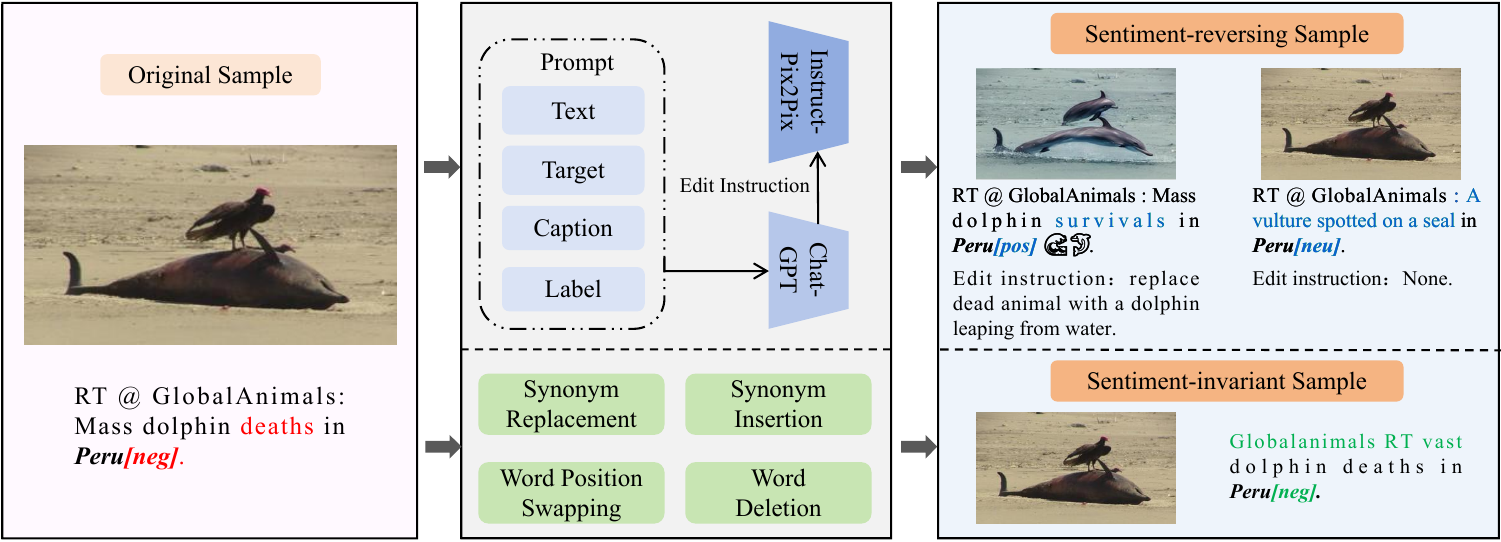} 
\caption{ Counterfactual data augmentation process with an example. When no image modification is required, the edit instruction will be set to ``none''.}
\label{fig2}
\end{figure}

We use counterfactual data to encourage the model to focus on learning key sentiment-related features by minimally editing specific causal features in the original samples. Two types of counterfactual samples are generated. The first generates samples with the opposite sentiment semantics but with similar contextual bias words. The second generates samples with the same sentiment semantics but with different contextual bias words. Figure~\ref{fig2} illustrates our data augmentation process.

\textbf{Sentiment-reversing Data Augmentation}. Since each data sample always contains at least one target word, with both textual and visual modalities reflecting its sentiment, we perform sample rewriting based on the target word for both the sentence and its associated image. We first edit the text modality. Since ChatGPT excels at understanding tasks and generating high-quality text tailored to specific goals and contexts~\cite{tan-etal-2024-large}, we use carefully designed instructions and constraints to guide it in generating counterfactual content. Each data sample is annotated with three sentiment labels: positive, negative, and neutral. For every sample, counterfactual examples corresponding to the other two sentiment labels need to be generated. When generating samples labeled as positive or negative, only the sentiment-related words or phrases linked to the target word are subtly adjusted, leaving the context and semantic information unrelated to the target word unchanged. This ensures the overall logic and theme of the text are preserved while reversing the sentiment expression of the target word. When generating samples labeled neutral, we focus on removing or replacing sentiment words or phrases around the target word with neutral and objective descriptions. By adopting factual statements and emphasizing the description of specific features or scenarios, the neutral samples retain the core meaning and syntactic structure of the original sentences while effectively reducing sentiment tendencies. In summary, regardless of whether positive, negative, or neutral samples are generated, the counterfactual text should maintain a similar word distribution to the original text, making only the necessary adjustments to sentiment features.

In addition to text modifications, localized adjustments to the associated image are essential to ensure consistency between the emotions conveyed by the text and the image across the pair data. Thus, we use an image editing model to make changes to the original image. Specifically, we utilize the InstructPix2Pix~\cite{brooks2023instructpix2pix} model, which combines the knowledge of large language models (GPT-3) and text-to-image models (Stable Diffusion), enabling the modification of specific objects or attributes within an image based on user instructions. We first convert the image into a detailed description $D$ and then input this description into ChatGPT to generate image editing instructions. To ensure that the counterfactual image aligns with the counterfactual sentiment in the text, we design the prompts to require the modification of the original image content to match the counterfactual sentiment. Additionally, to obtain concise and accurate image editing instructions, we standardize the format and length of the instructions, which may include adding or removing specific objects, adjusting facial expressions, modifying the background atmosphere, and so on. The prompt is input into ChatGPT:

\begin{align}
(S, T, D, y) \xrightarrow{\text{ChatGPT}} (\tilde{S},\tilde{y}, e),
\end{align}
where $\tilde{S}$ represents the counterfactual text,  $\tilde{y}$ represents the counterfactual label, and $e$ denotes the image editing instructions. Then, $e$ is sent to the image editing model:
\begin{align}
(V,e) \xrightarrow{\text{InstructPix2Pix}}\tilde{V},
\end{align}
where $\tilde{V}$ represents the counterfactual image.

\textbf{Sentiment-invariant Data Augmentation}. We utilize Easy Data Augmentation~\cite{wei2019eda} techniques to generate counterfactual samples with the same sentiment semantics but containing different biased words. First, some biased words in the sentence are randomly selected and replaced with their synonyms. Second, synonyms of the randomly selected biased words are inserted into the sentence. Third, two biased words are randomly selected and their positions are swapped. Finally, some biased words in the sentence are deleted with a certain probability. One of the above operations is randomly applied to produce the counterfactual sample $\{\overline{S}, V, T, y\}$.

\subsection{Contrastive Learning with Adaptive Weighting}

We introduce a contrastive learning mechanism to enhance the model's ability in perceiving sentiment consistency among positive samples and sentiment differences among negative samples. This mechanism pushes apart samples with different labels but similar word biases while bringing samples with the same label closer together. Our counterfactual data augmentation only modifies the key sentiment causal features while retaining most of the contextual words. This may cause the model to struggle to distinguish between different sentiment semantics effectively when the bias word similarity is high between samples with different labels. Therefore, we design an adaptive weighted contrastive strategy that encourages the model to focus more on samples with similar biased words but opposite labels. The contrastive loss is defined as follows: 
\begin{align}
\ \mathcal{L}_c = \frac{1}{|P|} \sum_{p \in P} -\log \frac{\exp \left( \psi(h) \cdot \psi(h_p) /  \tau  \right)}{\sum_{n \in N} w \cdot \exp\left( \psi(h) \cdot \psi(h_n) /  \tau  \right)},
\end{align}
where $P$ represents the set of positive samples that share the same label as the sample $x$, $N$ represents the set of negative samples, and $\tau$ is the temperature scaling parameter. $h_p$ and $h_n$ represent the final multimodal representations of the positive sample $p$ and negative sample $n$, respectively. $\psi$ is the MLP projection layer that transforms features, optimizing the space so similar samples are closer and dissimilar samples are farther apart. The weight $w$ is assigned to each sample pair based on the feature distance. Specifically, we first measure the biased similarity of two samples using the Euclidean Distance $d$. The distance $d(h, h_n)$ between $h$ of the sample $x$ and $h_n$ of the negative sample $n$ is calculated by:
\begin{align}
d(h, h_n) = \| h - h_n \|_2,
\end{align}
where $\| \|_2$ denotes the L2 norm. Based on the distance, the weight for $n$ is computed by:
\begin{align}
w =\exp(-d(h, h_n)).
\end{align}
The smaller the distance (i.e., the more similar the sample), the larger $w$ becomes.

\subsection{Model Specification}
To implement our proposed data augmentation and contrastive learning, we design a commonly used model architecture that consists of two main components: multi-modal feature extraction and fusion modules.

\textbf{Multi-modal Feature Extraction}. We encode the sentence, target word, and image using modality-specific encoders to extract multimodal features from the sample. For text representation, we use the pre-trained language model BERT as the text encoder and the target encoder to obtain the hidden representation of sentence $S$ and target $T$ respectively: $H_S=BERT(S), H_T = BERT(T)$, where $H_S \in\mathbb{R}^{n\times d}$, $H_T \in \mathbb{R}^{m\times d}$, $d$ is the hidden dimension, $n$ and $m$ are the length of $S$ and $T$. For image representation, we divide the input image into non-overlapping patches and feed them into a pre-trained image encoder ViT~\cite{dosovitskiy2020image}: $H_V=ViT(V) \in\mathbb{R}^{q\times d}$ containing the visual semantics, where $q$ is the number of non-overlapping image patches.

\textbf{Multi-modal Feature Fusion}. After obtaining the text representation $H_S$, target representation $H_T$, and image representation $H_V$, we adopt the widely used cross-attention mechanism~\cite{vaswani2017attention} to interact information among the three, extracting information related to the target word from both textual and visual modalities. We employ a multi-head attention mechanism to model the interaction between the target and text as well as the interaction between the target and image, where the target representation $H_T$ serves as query Q, and the text representation $H_S$ serves as keys $K$ and values $V$:
\begin{align}
H^i_{T{\rightarrow}S} &= softmax(\frac{H_T W^i_Q {W^i_K}^{\top} {H_S}^{\top} }{\sqrt{d_k}})\cdot{H_S W^i_V} ,\\
H_{T{\rightarrow}S}&=Concate(H^o_{T{\rightarrow}S}, H^1_{T{\rightarrow}S},..., H^a_{T{\rightarrow}S}),
\end{align}
where $H_{T{\rightarrow}S}$ is the generated target-based text representation, $W^i_Q\in \mathbb{R}^{d\times d_k}$, $W^i_K\in \mathbb{R}^{d\times d_k}$, and $W^i_V\in \mathbb{R}^{d\times d_k}$ are the learnable parameters. $d_k$ is the dimension of each attention head and $a$ is the number of attention heads. Following the above procedure, we could get the target-based image representation $H_{T{\rightarrow}V}$. Then, we concatenate $H_{T{\rightarrow}S}$ and $H_{T{\rightarrow}V}$ together to obtain the multimodal representation $H$. $H$ is first fed into a multi-head self-attention~\cite{tsai2019multimodal} as query, key, and value. It is subsequently passed through two layers of normalization (LN) and a feed-forward network (FFN) to obtain higher-level representations as follows:
\begin{align}
{Z} &= LN(H+MHSA(H,H,H)),\\
\tilde{H} &= LN(FFN(Z)+Z),
\end{align}
where $\tilde{H} \in \mathbb{R}^{m\times d}$ denotes the final multimodal representation. We feed the first token [CLS] of the $\tilde{H}$ to a softmax layer for the sentiment classification as follows:
\begin{align}
\hat{y} &= softmax(W^{\top}{{\tilde{H}_{[cls]}}}+b),
\end{align}
where $W \in \mathbb{R}^{d\times 3}$ and $b$ are learnable parameters.

\subsection{Objective Function}
Using both counterfactual and original data as model inputs, we apply the standard cross-entropy loss function to guide the model's classification as follows:
\begin{align}
 \mathcal{L}_s = - \frac{1}{|\hat{\mathcal {D}}|}\sum_{j=1}^{|\hat{\mathcal {D}}|}y_jlog\hat{y}_j,
\end{align}
where $\hat{\mathcal {D}}$ is the mixture of $\mathcal {D}$ and counterfactual samples. Our proposed contrastive loss is combined with the classification loss described above to form the final objective function for optimizing model parameters as follows:
\begin{align}
 \mathcal{L} = \mathcal{L}_s + \lambda {\mathcal{L}}_c,
\end{align}
where $\lambda$ controls the proportion of different loss functions.

\section{Experiment}

\subsection{Experimental Setups}
\textbf{Datasets}. To evaluate the effectiveness of our method, we conduct experiments on two benchmark datasets, including Twitter-2015 and Twitter-2017~\cite{yu2019adapting} which are collected multimodal tweets. Each dataset is divided into train, validation, and test sets. The former contains a total of 5338 samples, while the latter consists of 5972 samples.

\textbf{Implementation Details}. In the proposed framework, we adopt the pre-trained BERTweet~\cite{nguyen2020bertweet} as the target and text encoder to obtain the initial textual representations. VIT is employed as the image encoder to obtain the initial visual representations. The number of image patches is set to 16. We freeze all parameters of the image encoder ViT and use only its extracted features. We use GPT-4o and InstructPix2Pix for counterfactual data augmentation. Besides, we set the attention head, learning rate, warm-up rate, batch size, temperature scaling parameter, and dimension to 12, 2e-5, 0.1, 32, 0.07, and 768, respectively. Meanwhile, the hyper-parameter $\lambda$ is set to 0.8. We use the AdamW~\cite{loshchilov2019decoupled} optimizer to optimize the loss function. Both accuracy (Acc) and Macro-F1 score (F1) are used as evaluation metrics.

\textbf{Baselines}. We evaluate the performance of our proposed framework by comparing it with several competitive baselines across different modalities, including 1) the image-only method: ResNet-Target~\cite{he2016deep}, 2) the text-only method:
MemNet~\cite{tang2016aspect}, MGAN~\cite{fan2018multi}, and BERT~\cite{devlin2018bert}, and 3) the multimodal method: TomBERT~\cite{yu2019adapting}, EF-CapTrBERT~\cite{khan2021exploiting}, EF-CapTrBERT-DE, MPFIT~\cite{yang2024prompt}, FITE-DE~\cite{yang2022face}, AME-GCN~\cite{xiao2022adaptive}, VEMP~\cite{yang2023visual}, DPFN~\cite{wang2024dual}, A\textsuperscript{2}II~\cite{feng2024a2II}, and ITC-AOF~\cite{wang2024image}.

\subsection{Experimental Results}
The main results are shown in Table~\ref{tab2}. It is evident that our method performs best on both datasets compared to all unimodal (i.e., relying on texts or images) and multimodal methods (i.e., using both texts and images), demonstrating the validity of our proposed CED. Specifically, compared to text-only methods, ResNet-Target performs worse. This indicates that text-only methods can achieve relatively better performance while relying solely on images for prediction tends to yield poorer results.  The effective utilization of semantic information from text surpasses that from images, and text may contain more semantic information than images. Among the multimodal methods, on the Twitter-2015 dataset, our approach surpasses the strongest baseline, DPFN, by 0.9\% and 0.5\% in terms of Acc and F1 scores, respectively. Compared with TC-AOF on Twitter-2017, CED also outperforms it by 1.1\% in Acc and 1.2\% in F1. This demonstrates the effectiveness of our approach in removing word-level biases. Furthermore, Table \ref{tab3} presents the experimental results of applying our method to different baseline models, showing that the performance of these baseline models is improved. This not only proves the strong compatibility of our framework but also shows that CED, through counterfactual data augmentation and adaptive contrastive learning, can better capture sentiment-critical features related to target words within multimodal data, resulting in strong TMSC performance.

\begin{table}[t]
\centering
\caption{Results of different methods for two Twitter datasets. The mark $\dagger$ refers to $\texttt{p-value \textless 0.01}$ when comparing with the best baselines based on the significance t-test.} 
\resizebox{0.8\linewidth}{!}{
\begin{tabular}{lccccc}
\toprule
\multirow{2.5}{*}{Method} & \multicolumn{2}{c}{Twitter-2015} & & \multicolumn{2}{c}{Twitter-2017} \\
\cmidrule{2-3} \cmidrule{5-6}
&  Acc & F1 & &  Acc & F1\\
\toprule
&\multicolumn{5}{c}{Image}  \\
\cmidrule{2-6} 
ResNet-Target & 59.49 & 47.79 & &57.86  & 53.98\\
\toprule
& \multicolumn{5}{c}{\centering Text}  \\
\cmidrule{2-6}
MemNet & 70.11 &61.76 & & 64.18 & 60.90\\
MGAN & 71.17 &64.21 & & 64.75 & 61.46 \\
BERT &74.15  &68.86  &&68.15  &65.23\\
\toprule
&\multicolumn{5}{c}{Text + Image}  \\
\cmidrule{2-6} 
TomBERT &77.15 &71.75 & &70.34 &68.03 \\
EF-CapTrBERT &78.01 &73.25 & &69.77 &68.42\\
EF-CapTrBERT-DE &77.92 &73.90 & &72.30 &70.20\\
MPFIT &77.53 &73.53 &&70.35 &68.84\\
FITE-DE &78.76 &74.79 & &73.87 &73.03 \\
AME-GCN &79.75 &75.03 &&71.20 &69.59\\
VEMP &78.88 &75.09 &&73.01 &72.42 \\
DPFN  &79.9  &76.0  & &73.9  &72.6\\
A\textsuperscript{2}II &79.46 &75.16 &&74.39 &72.35\\
ITC-AOF & 79.45  &75.11  & &74.47 & 73.05\\
\toprule
CED &\textbf{80.84}$^\dagger$ & \textbf{76.53}$^\dagger$ & & \textbf{75.54}$^\dagger$ & \textbf{74.26}$^\dagger$ \\
\bottomrule
\end{tabular}
}
\label{tab2}
\end{table}

\begin{table}[t]
\centering
\caption{The effectiveness of CED on competitive baselines. The mark * refers to the results obtained by adding our method to the baseline model.} 
\resizebox{0.8\linewidth}{!}{
\begin{tabular}{lccccc}
\toprule
\multirow{2.5}{*}{Method} & \multicolumn{2}{c}{Twitter-2015} & & \multicolumn{2}{c}{Twitter-2017} \\
\cmidrule{2-3} \cmidrule{5-6}
&  Acc & F1 & &  Acc & F1\\
\midrule
ITC-AOF* &80.29	&75.94 && 75.65 & 74.81\\
FITE-DE*& 79.87 &75.71 & & 75.02 & 73.92\\
EF-CapTrBERT-DE*&79.43 & 74.63 && 74.08& 71.85\\
\bottomrule
\end{tabular}
}
\label{tab3}
\end{table}

\subsection{Ablation Study}
To verify the effectiveness of each component in CED, we conduct ablation experiments on two datasets, and the results are shown in Table~\ref{tab4}. Firstly, when sentiment-reversing counterfactual samples (w/o Senti-rev) are removed, leaving only original and sentiment-invariant counterfactual samples, the model's performance declines, suggesting these samples help capture sentiment features related to target words and enhance emotional understanding in multimodal contexts. Secondly, removing sentiment-invariant counterfactual samples (w/o Senti-inv) while retaining original samples and sentiment-reversing ones also lowers performance, indicating their role in improving sentiment feature recognition and generalization. Thirdly, when all counterfactual samples (w/o Counter-fact) are excluded but the adaptive contrastive learning mechanism is retained, performance drops further, highlighting the importance of counterfactual samples in enhancing sensitivity to target-word-related sentiments. Finally, removing the adaptive contrastive learning module (w/o Adapt-contra) worsens the model performance, as this mechanism helps the model learn robust emotional representations, reducing bias and improving sentiment classification accuracy.

\begin{table}[t]
\centering
\caption{Ablation Study.} 
\resizebox{0.7\linewidth}{!}{
\begin{tabular}{lccccc}
\toprule
\multirow{2.5}{*}{Method} & \multicolumn{2}{c}{Twitter-2015} & & \multicolumn{2}{c}{Twitter-2017} \\
\cmidrule{2-3} \cmidrule{5-6}
&  Acc & F1 & &  Acc & F1\\
\midrule
CED & \textbf{80.84} & \textbf{76.53} & & \textbf{75.54} & \textbf{74.26} \\
\midrule
w/o Senti-rev &79.33 & 75.42 && 74.49 & 73.26\\
w/o Senti-inv &79.93 & 76.02 && 74.88 & 73.74\\
w/o Counter-fact &79.13 & 75.05 &&74.32 & 72.79\\
w/o Adapt-contra &79.61 & 75.36 && 74.11 & 72.87\\
\bottomrule
\end{tabular}
}
\label{tab4}
\end{table}

\begin{figure}[t]
\centering 
\includegraphics[scale=0.49]{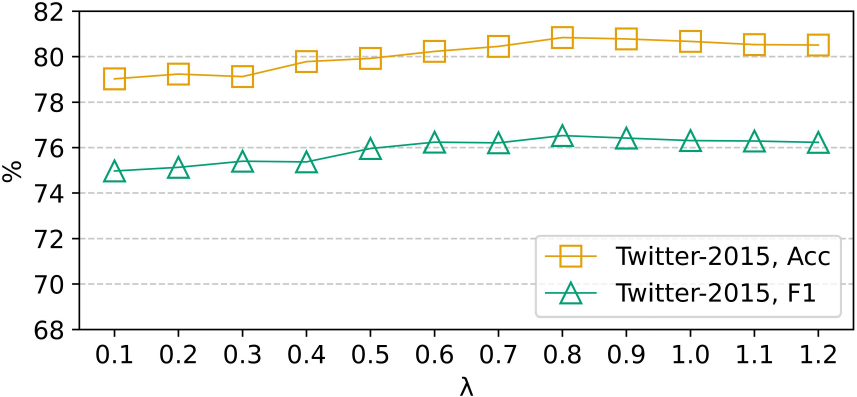} 
\caption{The effect of the hyper-parameter $\lambda$ on Twitter-2015 test sets.}
\label{fig3}
\end{figure}

\subsection{Effects of $\lambda$}
To investigate the effect of the hyper-parameter $\lambda$ on performance, we conduct experiments with the values of $\lambda$ ranging from 0 to 1.2 in increments of 0.1. Figure~\ref{fig3} presents the performance on Twitter-2015. The results indicate that as $\lambda$ increases, the performance of CED initially improves but levels off after reaching a certain value. The gradual integration of adaptive contrastive learning enables the model to optimize the feature space by more effectively separating samples with similar biased words but different labels while bringing samples with the same labels closer together. However, when the weight $\lambda$ of contrastive loss becomes too large, it diverts the model’s focus from the primary classification task, thereby limiting overall performance. According to the curve's trend, we set it to 0.8 on two  Twitter datasets.

\subsection{Case Study}
\begin{figure}[t]
\centering 
\includegraphics[scale=0.45]{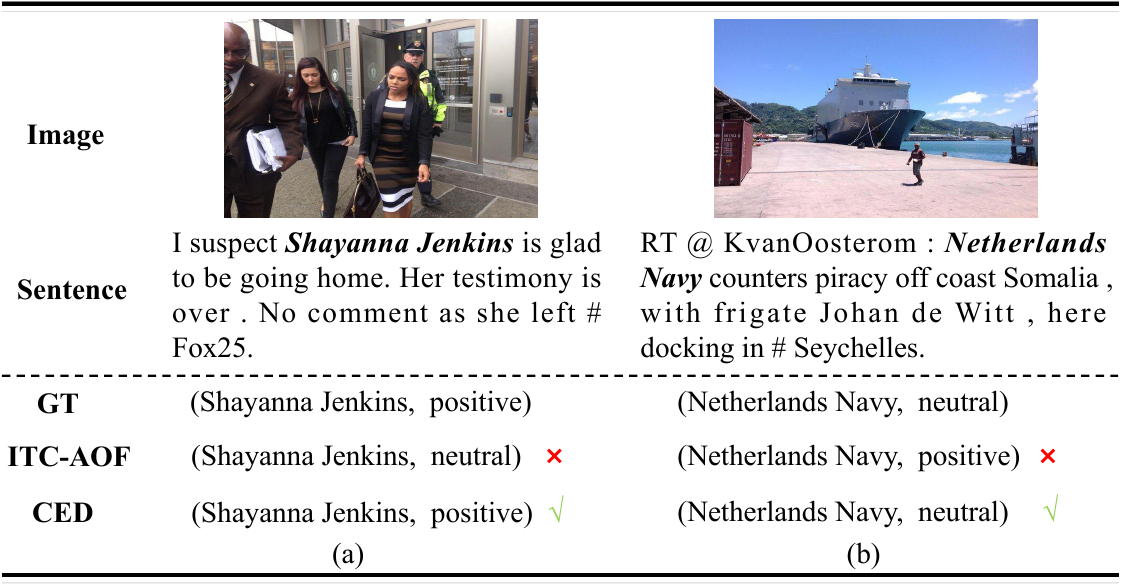} 
\caption{Case study. The symbols \ding{51} and \ding{53} represent correct and incorrect classification predictions, respectively.}
\label{fig4}
\end{figure}

We conduct a case study to further demonstrate the effectiveness of our method. As shown in Figure~\ref{fig4} (a), ITC-AOF makes incorrect predictions. The reason may be that its model is misled by biased words such as ``over'' and ``as'' in the sentence. These words have a high co-occurrence of the neutral label in the training samples, which may lead the model to focus on superficial associations between biased words and labels instead of capturing multimodal semantic content related to sentiment. In contrast, our CED leverages counterfactual data augmentation and adaptive debiased contrastive learning to obtain effective sentiment-related features, mitigating the issue of false correlations. A similar observation can be found in the case shown in Figure \ref{fig4} (b), where the model is misled by biased words such as ``off'' and ``here'' in the sentence, resulting in an incorrect prediction.

\section{Conclusion}
In this paper, we analyze the spurious correlations introduced by TMSC data, focusing on word-level biases between the text modality and output labels. To mitigate this, we propose a novel counterfactual-enhanced debiasing framework for TMSC. It employs a counterfactual multimodal data augmentation strategy to enhance the model’s sensitivity to sentiment-related features. Additionally, our adaptive contrastive learning mechanism enables the model to learn robust features from biased data. Experimental results on two benchmark datasets show that our method outperforms state-of-the-art baselines.

\section*{Acknowledgments}
This work was supported by the National Natural Science Foundation of China (62406081), and the Guangxi Natural Science Foundation (No. 2025GXNSFBA069232).

\bibliographystyle{IEEEbib}
\bibliography{icme2025references}

\begin{thebibliography}{10}

\bibitem{yu2019adapting}
Jianfei Yu and Jing Jiang,
\newblock ``Adapting bert for target-oriented multimodal sentiment classification,''
\newblock in {\em IJCAI}, 2019, pp. 5408--5414.

\bibitem{yang2024prompt}
Dan Yang, Xiuhong Li, Zhe Li, Chenyu Zhou, Xiaofan Wang, and Fan Chen,
\newblock ``Prompt fusion interaction transformer for aspect-based multimodal sentiment analysis,''
\newblock in {\em ICME}. IEEE, 2024, pp. 1--6.

\bibitem{khan2021exploiting}
Zaid Khan and Yun Fu,
\newblock ``Exploiting bert for multimodal target sentiment classification through input space translation,''
\newblock in {\em 29th ACM international conference on multimedia}, 2021, pp. 3034--3042.

\bibitem{wang2024image}
Qianlong Wang, Hongling Xu, Zhiyuan Wen, Bin Liang, Min Yang, Bing Qin, and Ruifeng Xu,
\newblock ``Image-to-text conversion and aspect-oriented filtration for multimodal aspect-based sentiment analysis,''
\newblock {\em IEEE Transactions on Affective Computing}, vol. 15, no. 3, pp. 1264--1278, 2024.

\bibitem{xiao2022adaptive}
L.~Xiao, E.~Zhou, X.~Wu, et~al.,
\newblock ``Adaptive multi-feature extraction graph convolutional networks for multimodal target sentiment analysis,''
\newblock in {\em ICME}. 2022, pp. 1--6, IEEE.

\bibitem{wang2024dual}
Di~Wang, Changning Tian, Xiao Liang, Lin Zhao, Lihuo He, and Quan Wang,
\newblock ``Dual-perspective fusion network for aspect-based multimodal sentiment analysis,''
\newblock {\em IEEE Transactions on Multimedia}, vol. 26, pp. 4028--4038, 2024.

\bibitem{wu2024diner}
Jialong Wu, Linhai Zhang, Deyu Zhou, and Guoqiang Xu,
\newblock ``{DINER}: Debiasing aspect-based sentiment analysis with multi-variable causal inference,''
\newblock in {\em Findings of the Association for Computational Linguistics: ACL 2024}. Aug. 2024, pp. 3504--3518, Association for Computational Linguistics.

\bibitem{ye2023rethinking}
J.~Ye, J.~Zhou, J.~Tian, et~al.,
\newblock ``Rethinking tmsc: An empirical study for target-oriented multimodal sentiment classification,''
\newblock in {\em Findings of ACL: EMNLP}, 2023, pp. 270--277.

\bibitem{devlin2018bert}
Jacob Devlin, Ming-Wei Chang, Kenton Lee, and Kristina Toutanova,
\newblock ``{BERT}: Pre-training of deep bidirectional transformers for language understanding,''
\newblock in {\em NAACL}, 2019, pp. 4171--4186.

\bibitem{he2016deep}
Kaiming He, Xiangyu Zhang, Shaoqing Ren, and Jian Sun,
\newblock ``Deep residual learning for image recognition,''
\newblock in {\em CVPR}, 2016, pp. 770--778.

\bibitem{yang2023visual}
B.~Yang and J.~Li,
\newblock ``Visual elements mining as prompts for instruction learning for target-oriented multimodal sentiment classification,''
\newblock in {\em EMNLP}, 2023, pp. 6062--6075.

\bibitem{feng2024a2II}
Junjia Feng, Mingqian Lin, Lin Shang, and Xiaoying Gao,
\newblock ``Autonomous aspect-image instruction a\textsuperscript{2}ii: Q-former guided multimodal sentiment classification,''
\newblock in {\em LREC-COLING}, 2024, pp. 1996--2005.

\bibitem{kaushik2020learning}
D.~Kaushik, E.~Hovy, and Z.~Lipton,
\newblock ``Learning the difference that makes a difference with counterfactually-augmented data,''
\newblock in {\em ICLR}, 2020.

\bibitem{chang2024counterfactual}
Mingshan Chang, Min Yang, Qingshan Jiang, and Ruifeng Xu,
\newblock ``Counterfactual-enhanced information bottleneck for aspect-based sentiment analysis,''
\newblock in {\em AAAI}, 2024, vol.~38, pp. 17736--17744.

\bibitem{wang2021robustness}
Zhao Wang and Aron Culotta,
\newblock ``Robustness to spurious correlations in text classification via automatically generated counterfactuals,''
\newblock in {\em AAAI}, 2021, vol.~35, pp. 14024--14031.

\bibitem{xin2024diffusion}
D.~Xin, J.~Yuan, and Y.~Li,
\newblock ``Diffusion based counterfactual augmentation for dual sentiment classification,''
\newblock in {\em LREC-COLING}, 2024, pp. 4901--4911.

\bibitem{jia2024debiasing}
Mengzhao Jia, Can Xie, and Liqiang Jing,
\newblock ``Debiasing multimodal sarcasm detection with contrastive learning,''
\newblock in {\em AAAI}, 2024, vol.~38, pp. 18354--18362.

\bibitem{tan-etal-2024-large}
Zhen Tan, Dawei Li, Song Wang, Alimohammad Beigi, Bohan Jiang, Amrita Bhattacharjee, Mansooreh Karami, Jundong Li, Lu~Cheng, and Huan Liu,
\newblock ``Large language models for data annotation and synthesis: A survey,''
\newblock in {\em EMNLP}, Nov. 2024, pp. 930--957.

\bibitem{brooks2023instructpix2pix}
T.~Brooks, A.~Holynski, and A.~A. Efros,
\newblock ``Instructpix2pix: Learning to follow image editing instructions,''
\newblock in {\em CVPR}, 2023, pp. 18392--18402.

\bibitem{wei2019eda}
J.~W. Wei and K.~Zou,
\newblock ``Eda: Easy data augmentation techniques for boosting performance on text classification tasks,''
\newblock in {\em EMNLP-IJCNLP}, 2019, pp. 6381--6387.

\bibitem{dosovitskiy2020image}
A.~Dosovitskiy, L.~Beyer, A.~Kolesnikov, D.~Weissenborn, X.~Zhai, T.~Unterthiner, M.~Dehghani, M.~Minderer, G.~Heigold, S.~Gelly, et~al.,
\newblock ``An image is worth 16x16 words: Transformers for image recognition at scale,''
\newblock {\em arXiv preprint arXiv:2010.11929}, 2020.

\bibitem{vaswani2017attention}
A.~Vaswani, N.~Shazeer, N.~Parmar, J.~Uszkoreit, L.~Jones, A.~N. Gomez, Ł. Kaiser, and I.~Polosukhin,
\newblock ``Attention is all you need,''
\newblock in {\em Advances in Neural Information Processing Systems}, 2017.

\bibitem{tsai2019multimodal}
Y.~H.~H. Tsai, S.~Bai, P.~P. Liang, J.~Z. Kolter, L.~P. Morency, and R.~Salakhutdinov,
\newblock ``Multimodal transformer for unaligned multimodal language sequences,''
\newblock in {\em ACL}, 2019, pp. 6558--6569.

\bibitem{nguyen2020bertweet}
Dat~Quoc Nguyen, Thanh Vu, and Anh-Tuan Nguyen,
\newblock ``Bertweet: A pre-trained language model for english tweets,''
\newblock in {\em EMNLP: System Demonstrations}, 2020, pp. 9--14.

\bibitem{loshchilov2019decoupled}
I.~Loshchilov and F.~Hutter,
\newblock ``Decoupled weight decay regularization,''
\newblock in {\em ICLR}, 2019.

\bibitem{tang2016aspect}
Duyu Tang, Bing Qin, and Ting Liu,
\newblock ``Aspect level sentiment classification with deep memory network,''
\newblock in {\em EMNLP}, 2016, pp. 214--224.

\bibitem{fan2018multi}
Feifan Fan, Yansong Feng, and Dongyan Zhao,
\newblock ``Multi-grained attention network for aspect-level sentiment classification,''
\newblock in {\em EMNLP}, 2018, pp. 3433--3442.

\bibitem{yang2022face}
Hao Yang, Yanyan Zhao, and Bing Qin,
\newblock ``Face-sensitive image-to-emotional-text cross-modal translation for multimodal aspect-based sentiment analysis,''
\newblock in {\em EMNLP}, 2022, pp. 3324--3335.

\end{thebibliography}

\vspace{12pt}
\end{document}